\documentclass[10pt,twocolumn,letterpaper]{article}

\usepackage{wacv}
\usepackage{times}
\usepackage{epsfig}
\usepackage{graphicx}
\usepackage{amsmath}
\usepackage{amssymb}
\usepackage{algorithm}
\usepackage{algorithmic}
\usepackage{multirow}
\usepackage{bm}

\wacvfinalcopy 

\ifwacvfinal
\fi

\ifwacvfinal
\usepackage[breaklinks=true,bookmarks=false]{hyperref}
\else
\usepackage[pagebackref=true,breaklinks=true,colorlinks,bookmarks=false]{hyperref}
\fi

\begin{document}

\title{Progressive Automatic Design of Search Space for One-Shot Neural Architecture Search}

\author{Xin Xia\footnotemark[1] 
	\qquad  Xuefeng Xiao\footnotemark[1]   
	\qquad Xing Wang\footnotemark[1]  
	\qquad Min Zheng \\
	ByteDance Inc.\\
	{\tt\small \{xiaxin.97, xiaoxuefeng.ailab, wangxing.613, zhengmin.666\}@bytedance.com}
}

\maketitle
\renewcommand{\thefootnote}{\fnsymbol{footnote}} 
\footnotetext[1]{Equal contribution.}

\maketitle

\ifwacvfinal
\thispagestyle{empty}
\fi


\begin{abstract}
	Neural Architecture Search (NAS) has attracted growing interest. To reduce the search cost, recent work has explored weight sharing across models and made major progress in One-Shot NAS. However, it has been observed that a model with higher one-shot model accuracy does not necessarily perform better when stand-alone trained. To address this issue, in this paper, we propose \textbf{P}rogressive \textbf{A}utomatic  \textbf{D}esign of search space, named PAD-NAS. Unlike previous approaches where the same operation search space is shared by all the layers in the supernet, we formulate a progressive search strategy based on operation pruning and build a layer-wise operation search space. In this way, PAD-NAS can automatically design the operations for each layer and  achieve a trade-off between search space quality and model diversity. During the search, we also take the hardware platform constraints into consideration for efficient neural network model deployment. Extensive experiments on ImageNet show that our method can achieve state-of-the-art performance.
\end{abstract}

\section{Introduction}
Neural Architecture Search (NAS) has received increasing attention in both industry and academia, and demonstrated much success in various computer vision tasks, such as image recognition \cite{guo2019single,liu2018darts,zoph2016neural,zoph2018learning}, object detection \cite{chen2019detnas,tan2019efficientdet}, and image segmentation \cite{chen2020fasterseg,liu2019auto,Nekrasov_2019_CVPR}. Early work \cite{zoph2016neural,zoph2018learning} is mainly built on top of reinforcement learning (RL) \cite{williams1992simple}, where tremendous amount of time is needed to evaluate candidate models by training them from scratch. In \cite{real2019regularized}, the authors use evolutionary algorithm (EA) and achieve comparable result to RL. Recently, more and more researchers have adopted weight sharing approaches \cite{brock2018smash,liu2018darts,enas,xie2018snas,zhang2018you} to reduce the computation.  They utilize an over-parameterized network, which is defined to subsume all architectures and needs to be trained only once.
\begin{figure}[H]
	\centering
	\includegraphics[scale=0.31]{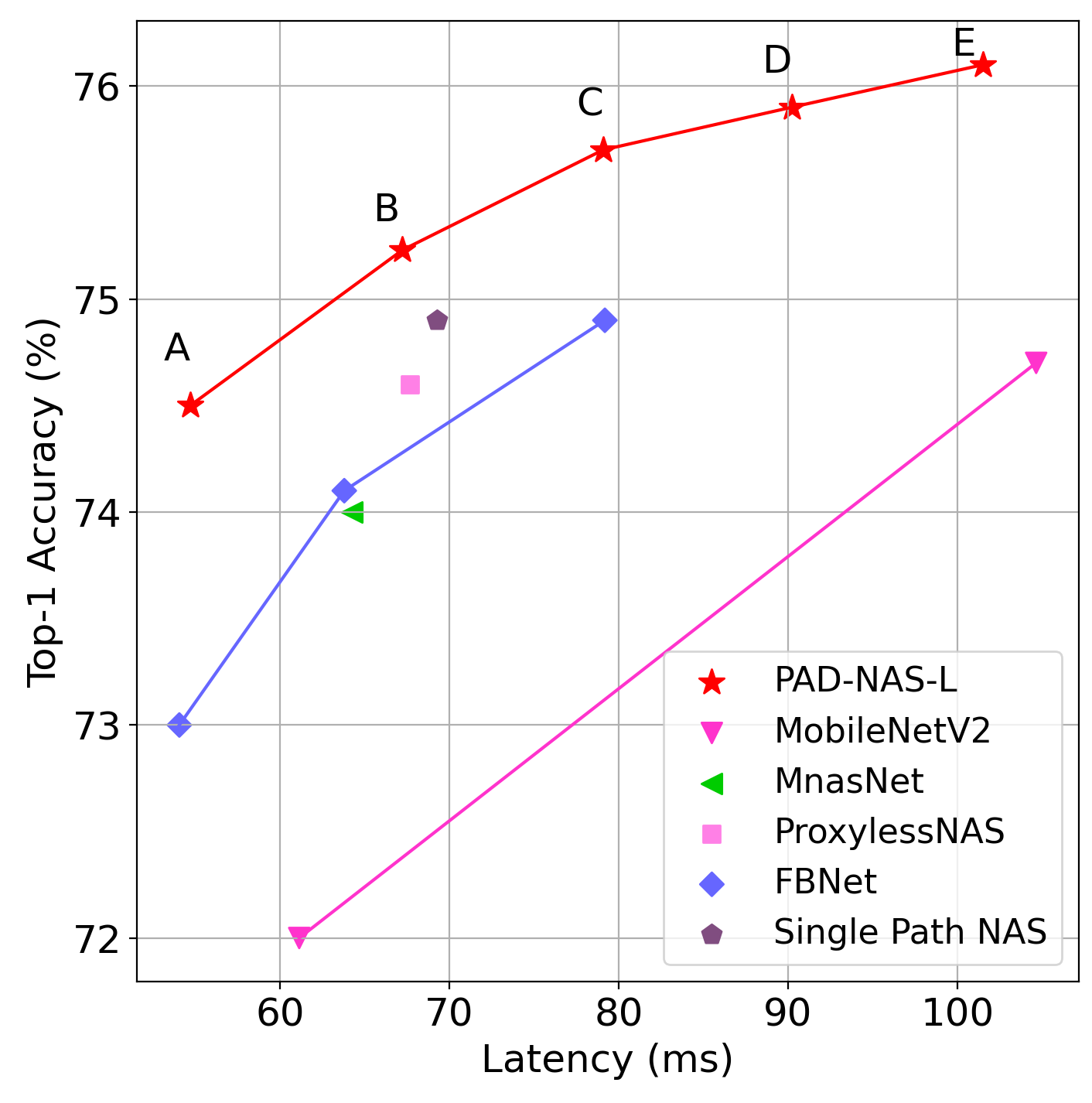}
	\caption{The trade-off between Pixel 3 latency and top-1 ImageNet accuracy. All models use the input resolution 224 and the latency is measured on a single large core of the same device using TFLite\cite{abadi2016tensorflow}.}
	\label{acc}
	\vspace{-0.3cm}
\end{figure}

Weight-sharing approaches can be mainly divided into two categories. In the first category, researchers use a continuous relaxation of search space, e.g. ProxylessNAS \cite{cai2018proxylessnas}, DARTS \cite{liu2018darts}, and FBNet \cite{wu2019fbnet}. The architecture distribution is continuously parameterized. Supernet training and architecture search are deeply coupled into single stage and jointly optimized by gradient based methods. Deep coupling between the architecture parameters and supernet weights introduces bias and instability to the search process. One-Shot NAS, e.g., SPOS \cite{guo2019single}, SMASH \cite{brock2018smash} and others \cite{bender2019understanding, chu2019fairnas}, belongs to the other category. The optimization of the supernet weights and architecture parameters are decoupled into two sequential steps. The fairness among all architectures is ensured by sampling architecture or dropping out operators uniformly. During the architecture search, the validation accuracy of a model is predicted by inheriting the weights from the trained supernet. Unfortunately, as stated in \cite{adam2019understanding,bender2019understanding,chu2019fairnas}, weight coupling issue exists in one-shot methods that one-shot model accuracy can not truly reflect the relative performance of architectures. 

In this paper, we aim to mitigate the negative effect of weight coupling in one-shot approaches from the view of operation search space.
Current one-shot approaches \cite{bender2019understanding,brock2018smash,chu2019fairnas,guo2019single} use the same operation search space for all the layers. However, in practice, we observe some operations will never be selected by certain layers in the final architectures. 
The reason lies in that sub-networks that contain these redundant operations either violate the hardware platform constraints, or perform poorly on the validation dataset and are excluded during the architecture search step. So a natural question arises: if an operation will never be selected, why spend the effort on training it at the very beginning? Keeping these operations will degrade the performance, since the more operations in the supernet, the more severe the interference between the operations. Thus, an effective way to mitigate the weight coupling is to remove these operations before training. A follow-up question is how we can identify if an operations is redundant or not, before the training of a supernet.

\begin{figure*}
	\vspace{-0.3cm}
	\centering
	\includegraphics[scale=0.25]{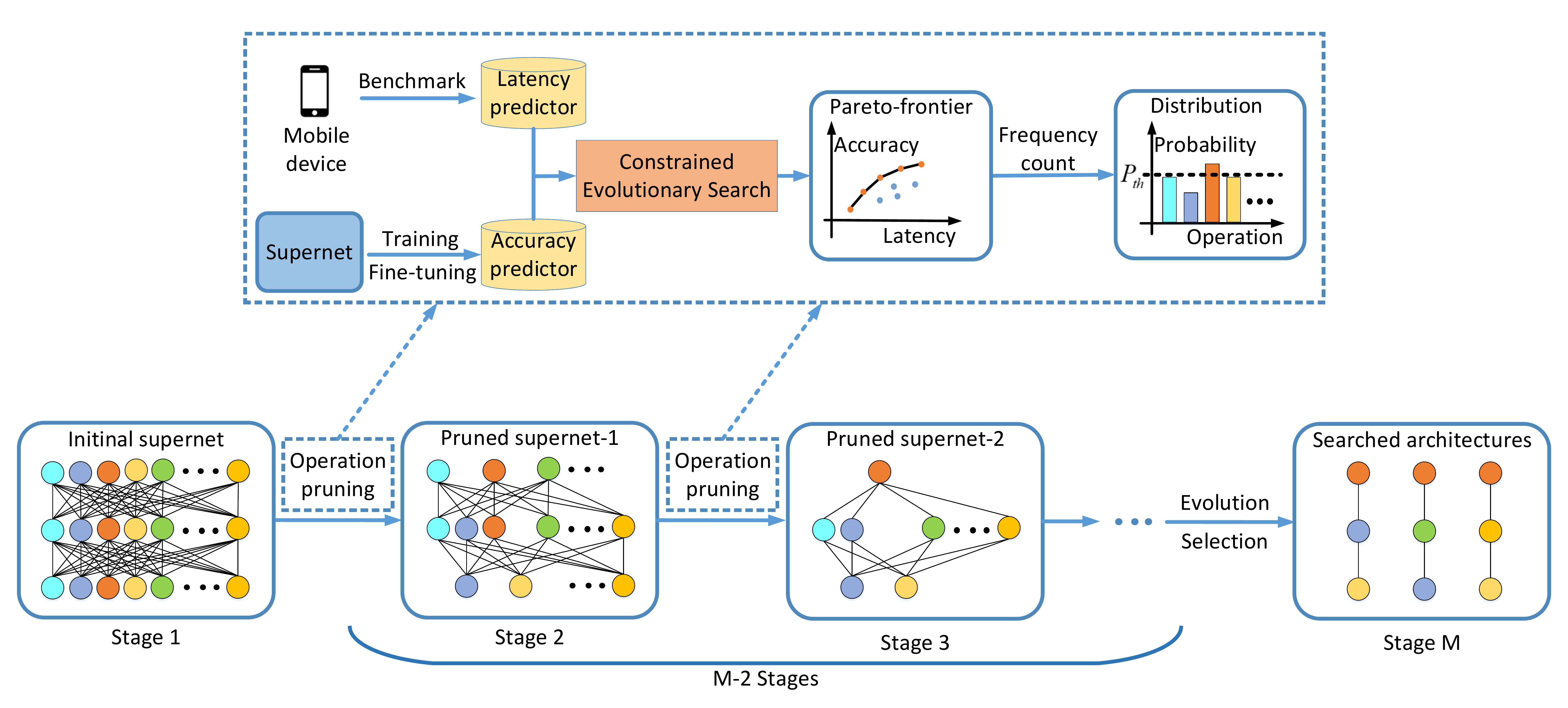}
	\caption{The overall framework of PAD-NAS. \textbf{Bottom:} PAD-NAS is divided into $M$ stages: the first stage is initial supernet training, and the next $M-$2 stages repeat the process of pruning the operations and training the pruned supernet. The last stage is the architecture evolution and selection. \textbf{Top:} The procedures of operation pruning. Supernet is used as an accuracy predictor after training, and a latency predictor is built for target mobile devices. The constrained evolutionary search algorithm is used to evolve the network architecture, and the probability distribution of the operations for each layer is estimated from Pareto frontier.}
	\label{framework}
	\vspace{-0.3cm}
\end{figure*}


To answer the questions above, we propose a simple yet effective approach named PAD-NAS. Our algorithm can automatically design its own operations for each layer and build a layer-wise search space through a progressive search strategy. The flow of our algorithm is illustrated in Fig.~\ref{framework}. In the first stage, we perform the training of an initial supernet, where the same operation search space is shared by all the layers. For the next $M-$2 stages, we start from the supernet coming from the previous stage, and estimate the operation probability distribution for each layer  from the architectures that reside on the Pareto frontier \cite{fleischer2003measure}. Next, we prune the operations layer by layer and remove the ones whose probabilities are below certain threshold to build the pruned supernet for the next stage. Finally, we finetune the pruned supernet. We repeat the process above until stopping criterion is satisfied. In the final stage, we search the architectures from the supernet and return the architectures with highest accuracy.


The effectiveness of PAD-NAS is demonstrated on ImageNet. 
We name the models discovered by PAD-NAS as PAD-NASs. 
PAD-NASs achieve state-of-the-art (SOTA) performance on ImageNet and outperform efficient networks designed manually and automatically, such as MobileNetV2 \cite{sandler2018mobilenetv2}, FBNet \cite{wu2019fbnet}, ProxylessNAS \cite{cai2018proxylessnas} and SPOS \cite{guo2019single}. As shown in Fig.~\ref{acc}, PAD-NAS-L-A achieves ${74.5\%}$ top-1 accuracy with 271M FLOPs and 54.7 ms latency on an Pixel 3 phone, ${1.5\%}$ higher than FBNet-A \cite{wu2019fbnet}. Top-1 accuracy of PAD-NAS-L-B is ${0.6\%}$ higher than Proxyless-R \cite{cai2018proxylessnas}, while PAD-NAS-L-C achieves ${0.8\%}$ absolute gain in top-1 accuracy compared with FBNet-C \cite{wu2019fbnet}. 

Our main contributions are summarized as follows:
\begin{itemize}
	\item 
	We present PAD-NAS, an efficient neural architecture search framework for One-Shot NAS. 
	It automatically designs the search space for each layer through a progressive search strategy.
	\item 
	we propose a new search space pruning method for One-Shot NAS, which can achieve a trade-off
	between search space quality and model diversity. 
	
	\item 
	Extensive experiments demonstrate the advantage of PAD-NAS. It achieves SOTA performance on ImageNet and significantly mitigates weight coupling.
\end{itemize}

\section{Related Work}
Recently, in order to reduce the computation cost of NAS, some researchers propose weight sharing approaches to speed up the architecture search, such as ENAS \cite{enas}, DARTS \cite{liu2018darts}, and One-Shot NAS \cite{bender2019understanding,guo2019single}. All sub-network architectures inherit the weights from the trained supernet without training from scratch. DARTS softens the discrete search space into a continuous search space and directly optimizes it by the gradient method. \cite{brock2018smash} and \cite{bender2019understanding} proposes a method which decouples supernet training and architecture search into two sequential stages, including supernet training and architecture search. However, due to the weight coupling in the supernet, the accuracy of the supernet prediction has a certain deviation from the ground truth, which result in inaccurate ranking of the architectures. The authors of SPOS \cite{guo2019single} further propose uniform sampling and single-path training to overcome weight coupling in One-Shot NAS. ProxylessNAS \cite{cai2018proxylessnas} binarize entire paths and keep only one path when training the over-parameterized supernet to reduce memory footprint. FBNet \cite{wu2019fbnet} use a proxy dataset (subset of ImageNet) to  train the continuously parameterized architecture distribution. FairNas \cite{chu2019fairnas} is based on \cite{guo2019single} and proposes a new fairness sampling and training strategy for supernet training. GreedyNAS \cite{you2020greedynas} propose to greedily focus on training potentially-good paths, which implemented by multi-path sampling strategy with rejection. ABS \cite{hu2020angle} propose an angle-based search space shrinking method by adopting a novel angle-based metric to evaluate capability of child models and guide the shrinking procedure. RegNet \cite{radosavovic2020designing} propose to design network design spaces, which parametrize populations of networks, and present a new network design paradigm.

Different from the methods above, we propose a progressive search strategy to reduce the number of operations in each stage and build a layer-wise operation search space for One-Shot NAS automatically. Our work is also closely related to P-DARTS\cite{Chen_2019_ICCV} and HM-NAS\cite{yan2019hm}. \cite{Chen_2019_ICCV} propose a progressive version of DARTS to bridge the depth gap between search and evaluation scenarios. Its core idea is to gradually increase the depth of candidate architectures. \cite{yan2019hm} incorporates a multi-level architecture encoding scheme to enable an architecture candidate to have arbitrary numbers of edges and operations with different importance. \cite{Chen_2019_ICCV, yan2019hm} belong to differential NAS, which prune the operations according to the value of architecture parameters. However, there are no architecture parameters in One-Shot NAS. 

Another relevant topic is network pruning \cite{han2015deep,He_2017_ICCV,Liu_2017_ICCV} that aim to reduce the network complexity by removing redundant, non-informative connections in a pre-trained network. Similar to this work, we start from an over-parameterized supernet and then prune the redundant operations to get the optimized architecture. The distinction is that they aim to prune the connections in a pre-trained network, while we focus on improving One-Shot NAS performance through operation pruning. 

\section{Method}
\subsection{Problem Formulation and Motivation}
One-Shot NAS \cite{bender2019understanding,brock2018smash} contains two stages. 
The first stage is supernet training, which is formulated as:
\begin{equation}
{W_S} = \mathop {\arg \min }\limits_W {Loss_{train}}(N(S,W)),
\label{eq1}
\end{equation}
where ${Loss_{train}(\cdot)}$ is the loss function on the training set,  ${N(S,W)}$ is the supernet, represented by the search space ${S}$ and its weights ${W}$, and ${W_S}$ are the learned weights of the supernet. 

The second stage is the architecture search. It aims to find the architectures from the supernet that has the best one-shot accuracy on the validation set under mobile latency constraint, expressed as:

\begin{equation}
\begin{split}
{s^*} &=  \mathop {\arg \max }\limits_{s \in S} Ac{c_{val}}(N(s,{W_S}(s)))\\
&s.t. {{\mathop{\rm Lat}\nolimits} _{\min }} \le { \mathop{\rm Latency}\nolimits} ({s^ * }) \le {{\mathop{\rm Lat}\nolimits} _{\max }},
\end{split}
\label{searchconstraints}
\end{equation}
where ${{\mathop{\rm Lat}\nolimits} _{\min}}$ and ${{\mathop{\rm Lat}\nolimits} _{\max}}$ are lower and upper mobile latency constraint. Each sampled sub-network inherits its weights from ${W_S}$ as ${W_S(s)}$. Therefore, one-shot accuracy ${Acc_{val}(\cdot)}$ only requires inference on validation dataset. This completes the search phase. In the evaluation phase, without violating the mobile latency constraint, the architectures with the highest one-shot accuracy will be selected to do the stand-alone training from scratch.

However, as stated by many researchers \cite{bender2019understanding,brock2018smash,guo2019single}, weight coupling issue exists in One-Shot NAS, which results in ranking inconsistency between supernet predicted accuracies and that of ground-truth ones by stand-alone training from scratch. Architectures with a higher supernet predicted accuracy during the search phase may perform worse in the evaluation phase. 

\begin{figure}
	\centering
	\includegraphics[scale=0.5]{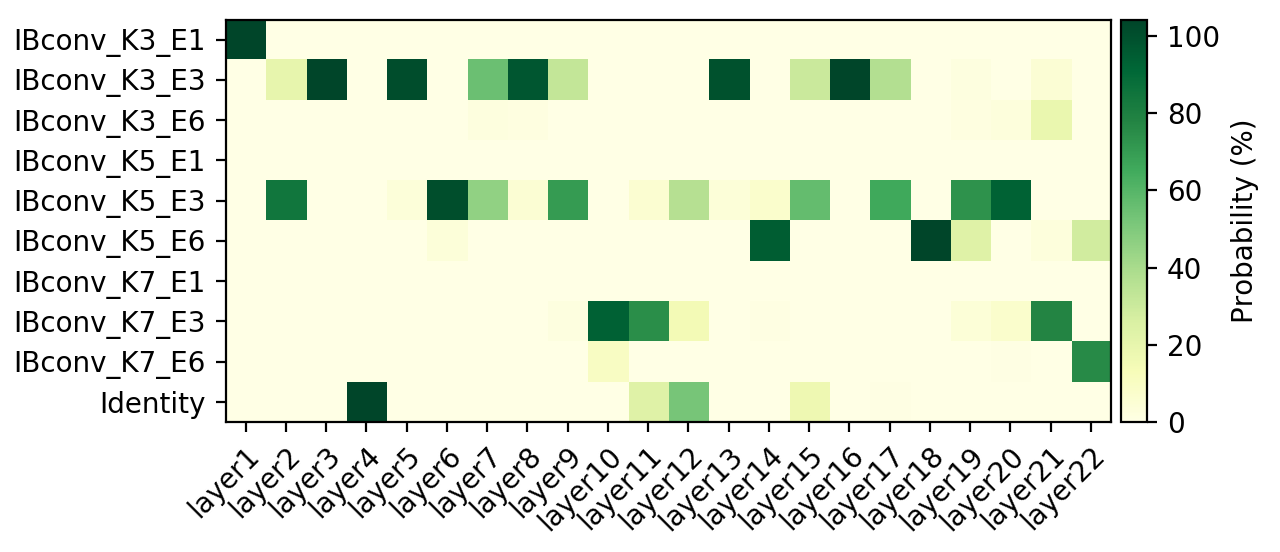}
	\caption{Probability distribution of each layer's operation in the searched architectures.}
	\label{opdistribution}
	\vspace{-0.6cm}
\end{figure}

In this paper, we aim to mitigate weight coupling and improve the performance of One-Shot NAS. We start with the analysis of operation search space. In practice, we observe that redundant operations exist in the supernet.
These redundant operations will never be selected by the searched architectures. As shown in Fig.~\ref{opdistribution}, where x-axis represents the layer name and y-axis represents the operation name, the operation distribution in each layer is sparse. For example, in layer1, the distribution concentrates on the operation IBconv\_K3\_E1, while all the other operations never appear in this layer. The reason lies in the architectures that contain the redundant operations either violate the constrains in Eq.~\ref{searchconstraints}, or has a low validation accuracy and are excluded during the architecture search. 

Motivated by \cite{han2015deep} where uncritical connections in deep networks can be removed without affecting the performance, we claim that existing operation search space is redundant, and needs to be pruned. However, different from network pruning \cite{han2015deep}, whose main purpose is to decrease the network complexity, our main goal is to mitigate weight coupling and improve One-Shot NAS performance by pruning unnecessary operations. These operations will hurt the supernet training since more operations means more intense interference between the operations in the supernet, resulting in more severe ranking inconsistency between supernet predicted accuracy and ground-truth one. However, large number of operations is demanded for NAS to discover promising models. Due to this conflict, we present PAD-NAS which follows a coarse-to-fine manner to refine the operation search by progressively pruning the operations and build the layer-wise operation search space automatically.

\subsection{Progressive Automatic Design of Search Space}
The whole process of PAD-NAS is elaborated in Algorithm~\ref{algor}. $s^{i, j}$ denotes the $i$th operation for layer $j$ in the supernet, $p^{i, j}$ represents the probability of operation $s^{i,j}$. 
We start with an initial operation search space, which is shared by all the layers in the supernet. Next, we construct the supernet following Table~\ref{searchspace} with this initial search space and train it with path-wise manner proposed in \cite{guo2019single}. This is the first stage of PAD-NAS, served as the initialization.  For the next $M-$2 stages, we first use constrained evolutionary algorithm to search the architectures that belong to Pareto frontier on the input supernet. The output supernet of the previous stage is fed as the input of the current stage.
Then, we estimate the probability distribution of each operation layer-by-layer from Pareto frontier and remove the operations whose probabilities are below certain threshold $P_{th}$. To this end, we build the layer-wise operation search space for the current stage. The last part is to finetune the pruned supernet and feed it into the next stage. The process above is repeated until we reach the targeted number of search stages. When it comes to the last stage of PAD-NAS, constrained evolutionary algorithm is used to search the architectures on the supernet, which is the output of the ($M-$1)th stage, and return the architectures with highest accuracy. We list more details as follows.

\begin{algorithm}[tb]
	\caption{PAD-NAS}
	\textbf{Input}: operation search space $S = \left \{ s^{i, j} \right \}$, probability threshold $P_{th}$, the number of progressive search stages ${M}$, latency constraints $Lat_{\min}$ and $Lat_{\max}$, and the number of layers $J$ in the supernet
	\begin{algorithmic}[1] 
		\STATE Initial supernet training: construct and train supernet $N(S, W)$
		\FOR {$k = 2$ to $M-1$}
		
		\STATE Architecture search: search the architectures on the supernet $N(S, W)$
		\FOR {$j = 1$ to $J$}
		\STATE Distribution estimation: count and normalize the frequency of $s^{i, j}$ to get $p^{i, j}$
		\ENDFOR
		\STATE Pruning: remove $s^{i, j}$ from $S$ if $p^{i, j} \leq P_{th} $ 
		\STATE Pruned supernet training: construct and finetune supernet $N(S, W)$
		\ENDFOR
		\STATE Repeat step 3
		\STATE \textbf{return} the architectures with highest accuracy
	\end{algorithmic}
	\label{algor}
\end{algorithm}

\noindent\textbf{Initial Supernet Training.} For the training of initial supernet, we adopt a path-wise \cite{guo2019single} manner, where supernet training and architecture search are decoupled into two sequential steps, to ensure the trained supernet can reflect the relative performance of sub-networks. However, since weight coupling is inevitable in weight-sharing approaches \cite{guo2019single,liu2018darts}, supernet predicted accuracy can only coarsely indicate the relative ranking of sub-networks.  The trained initial supernet is the starting point of PAD-NAS and 
served as the initialization. 

\noindent\textbf{Constrained Evolutionary Search.} After initial supernet training, we need to search the architectures. Evolutionary search performs better than random search in previous one-shot works \cite{bender2019understanding,brock2018smash,guo2019single}. The evolutionary algorithm is flexible in dealing with hardware platform constraints in Eq. \ref{searchconstraints}, since the mutation and
crossover processes can be manipulated to generate proper candidates to satisfy the constraints. We aim to estimate the probability distribution of operations by counting their frequencies in the searched architectures. The estimated operation distribution is unstable and exhibits a large variance if the evolutionary search algorithm in \cite{guo2019single} is used. 

To deal with this problem, we propose Constrained Evolutionary Search (CES), which is built on top of Non-Dominated Sorting Genetic Algorithm \uppercase\expandafter{\romannumeral2} (NSGA-\uppercase\expandafter{\romannumeral2}) \cite{deb2002fast}. Note that the authors of \cite{lu2019nsga} are the first to introduce NSGA-\uppercase\expandafter{\romannumeral2} into NAS.
Architecture search can be formulated as a multi-objective optimization problem to balance between mobile latency and architecture accuracy. We add latency constraint into the iteration of NSGA-\uppercase\expandafter{\romannumeral2} to formulate CES. More specifically, we discard the architectures that violate the constraint in the crossover and mutation of CES. This change gives us better result. Different from the evolutionary search in \cite{guo2019single} that architectures with poor accuracy are discarded in each iteration, no architectures are discarded but all of them are sorted by its latency and accuracy in each iteration. The estimated operation distribution by CES is stable and its corresponding variance is small.

During the search, the architecture accuracy is predicted by the supernet through weight inheritance, and its corresponding latency on mobile devices is estimated by the latency predictor. We build a lookup table of mobile latency for operations in the supernet and the architecture latency is predicted by summing up the latency of all of its operations.

\noindent\textbf{Operation Pruning.} We prune the operations based on its corresponding probability distribution. Since there is no way to get the ground-truth distribution, the sampling method is applied here. We count the frequency of operations in the architectures after architecture search and normalize it to get the approximate probability distribution. However, not all the architectures are equally important. Only the architectures that belong to Pareto frontier are used. These architectures dominate all the other ones. Here, we say one architecture dominates the others if and only if its latency is no bigger than others while its accuracy is no lower than others. Next, we remove the operations whose probability are below pre-defined threshold.

During CES, the architecture accuracy is predicted by supernet, which introduces error into the sorting part of CES, due to weight coupling. This implies Pareto frontier returned by CES is noisy, which may cause inaccurate estimation of operation probability distribution. To mitigate this effect, we count the frequency of operations from the architectures whose nondomination rank \cite{deb2002fast} is smaller than 10, where the nondomination rank of Pareto frontier is 1.


\noindent\textbf{Pruned Supernet Training.} After operation pruning, we get the pruned supernet. 
The training of pruned supernet is the same as initial supernet training. The only distinction is weights are randomly initialized in initial supernet training, while weights in the pruned supernet are inherited from the supernet in the previous stage. We only need to fine-tune the pruned supernet instead of training it from scratch. This reduces the search cost without affecting the performance.

\section{Experiments}
\subsection{Dataset and Implement Details}
\vspace{0.01cm}
\noindent\textbf{Dataset.} Throughout the paper, we use the ILSVRC2012 dataset \cite{Deng2009ImageNet}. To be consistent with previous works, 50 images are randomly sampled from each class of the training set, and a total of 50,000 images are used as the validation set. The original validation set is used as the test set.


\noindent\textbf{Search Space.} The whole network structure is shown in Table~\ref{searchspace}, which is used in previous work \cite{cai2018proxylessnas}. For fair comparison, our basic search space is the same as \cite{cai2018proxylessnas} and shown in the left column of Table~\ref{operation}. The optional operations in each search block structure (SBS) are 6 types of operation (with a kernel size 3$\times$3 , 5$\times$5 or 7$\times$7 and an expansion factor of 3 or 6), plus one identity operation. In addition, the expansion factor in the first SBS block is fixed as 1, and the identity operation is forbidden in the first layer of every block. The basic search space size is ${3 \times 6^6 \times 7^{15} \approx 6.64 \times 10^{17} }$. 
\setlength{\tabcolsep}{1pt}
\begin{table}[h]
		\vspace{-0.2cm}
	\centering
	\caption{Architecture of the supernet. Output channels denotes the output channel number of a block. Repeat denotes the number of blocks in a stage. Stride denotes the stride of the first block in a stage. The output channel size of the first three stage is 32-16-32 in basic search space and 16-16-24 in large search space.}
	\scalebox{0.9}{\begin{tabular}{lcccc}  
			\hline
			Input shape  & Block & Output channels   & Repeat & Stride  \\
			\hline
			\(224^2 \times 3\)             & \( 3 \times 3 \) conv  & 32 (16)   &1   & 2      \\
			\(112^2 \times 32 (16)\)             & SBS                         & 16 (16)   &1   & 1     \\
			\(112^2 \times 16 (16)\)             & SBS                          & 32 (24)  &4   & 2   \\
			\(56^2 \times 32 (24)\)              &SBS                          & 40         &4   & 2   \\
			\(28^2 \times 40\)              &SBS                          & 80        &4   & 2      \\
			\(14^2 \times 80\)              &SBS                         & 96          &4    & 1       \\
			\(14^2 \times 96\)              &SBS                         & 192        &4   & 2       \\
			\(7^2 \times 192\)              &SBS                          & 320       &1   & 1       \\
			\(7^2 \times 320\)             &\( 1 \times 1 \) conv   & 1280        &1   & 1       \\
			\(7^2 \times 320\)               &global avgpool         & -              &1   & 1       \\
			\(7^2 \times 1280\)             &fc                                & 1000  &1   & -       \\
			\hline
	\end{tabular}}
	\label{searchspace}
\end{table}
\setlength{\tabcolsep}{1.4pt}

\setlength{\tabcolsep}{4pt}
\begin{table}[h]
	\vspace{-0.5cm}
	\centering
	\caption{Operations table. IBConv\_KX\_EY represents the specific operator IBConv with expansion Y and kernel size X. IBConv denotes Inverted Bottleneck in MobilenetV2.}
	\scalebox{0.87}{\begin{tabular}{c|cc}  
			\hline
			\multirow{2}{*}{Basic search space}   & \multicolumn{2}{c}{Operators exclusively in } \\ 
			& \multicolumn{2}{c}{large search space} \\ 
			\hline
			IBConv\_K3\_E3& IBConv\_K3\_E1          &IBConv\_K5\_E4        \\
			IBConv\_K3\_E6&  IBConv\_K3\_E2          &IBConv\_K5\_E5         \\
			IBConv\_K5\_E3&  IBConv\_K3\_E4         &IBConv\_K7\_E1         \\
			IBConv\_K5\_E6&  IBConv\_K3\_E5          &IBConv\_K7\_E2        \\
			IBConv\_K7\_E3&  IBConv\_K5\_E1          &IBConv\_K7\_E4       \\
			IBConv\_K7\_E6&  IBConv\_K5\_E2          &IBConv\_K7\_E5        \\ 
			Identity      &                                           \\
			\hline
	\end{tabular}}
	\label{operation}
	\vspace{-0.3cm}
\end{table}

\setlength{\tabcolsep}{1.4pt}

Moreover, we enlarge the search space by adding 12 more operations and formulate the large search space, as shown in the right column of right column of Table~\ref{operation}. We do not add any special operations, but a fine-grained version of the basic search space. In the basic search space, the expansion is either 3 or 6, while in the large search space, the expansion ranges from 1 to 6.
The large search space size is ${3 \times 18^6 \times 19^{15} \approx 1.55 \times 10^{27} }$.
The primary reason to use the enlarged search space in this paper is to show PAD-NAS can automatically design the search space, mitigate the weight coupling issue, and bring better results, compared to the small search space. 


\noindent\textbf{Implementation Details.} For the stand-alone training of the searched architecture (after evolutionary search) from scratch, we use the same settings as \cite{guo2019single}. The network weights are optimized by momentum SGD, with an initial learning rate of 0.5, a momentum of 0.9, and a weight decay of \(4 \times 10^{-5}\).  A linear learning rate decay strategy is applied for 240 epochs and the batch size is 1024. 
It is worth noting that we have not used neither squeeze-and-excitation \cite{hu2018squeeze} nor swish activation functions \cite{ramachandran2017searching}, which are some tricks, not related to the weight coupling problem in this paper. Extra data augmentations such as mixup \cite{zhang2017mixup} and autoaugment \cite{Cubuk_2019_CVPR} are not used as well.

For the training of the supernet, we first train the initial supernet for 120 epochs with an initial learning rate of 0.5. All other training settings are the same as stand-alone training. Next, we apply operation pruning for the first time and finetune the pruned supernet for 80 epochs with an initial learning rate of 0.1. Finally, we prune the operations for the second time and finetune the pruned supernet-2 for another 40 epochs with an initial learning rate of 0.1. The total number of supernet training epochs is 240, exactly the same as stand-alone training.

We set the probability threshold ${P_{th}}$ in a heuristic way. We aim to balance between search space quality and model diversity. If $P_{th}$ is close to 1, most operations will be removed and model diversity is lost. If $P_{th}$ is close to 0, most operations will be kept and search space quality is degraded, since redundant operations exist. We observe that $1\%$ can achieve a good balance in not only MobilenetV2 space shown in Table~\ref{result}, but also in MobilenetV3 space presented in Table~\ref{mv3}.
And, the latency lower ${{\mathop{\rm Lat}\nolimits} _{min}}$ and upper bound  ${{\mathop{\rm Lat}\nolimits} _{max}}$ is set to 60ms and 70ms for basic search space and 50ms and 100ms for large search space. The latency is measured on Pixel 3 using TFLite. The number of progressive search stages ${M}$ is set to $4$.  The result  of $M \textgreater 4$ in our search space become saturated and is similar to $M = 4 $. In the constrained evolution search, the population size is set to 64, the evolution iteration is set to 40, and the polynomial mutation and two-point crossover is adopted.


\noindent\textbf{Notations.} Two baseline methods are frequently mentioned in the following part.  In the first baseline method, we fully train the initial supernet with single path and uniform sampling in \cite{guo2019single} and search the architectures with CES proposed in this paper. We name it I-Supernet (Initial Supernet). I-Supernet is a special case of PAD-NAS with $M=2$ in Fig.~\ref{framework} and Algorithm~\ref{algor}. There is no operation pruning inside. The second baseline method is called P-Supernet (Pruned Supernet).
P-Supernet corresponds to PAD-NAS with $M=3$, where we only prune the operations once. 
PAD-NAS denotes our main result, corresponding to PAD-NAS with $M=4$, where we prune the operations twice.
I-Supernet-S, P-Supernet-S, and PAD-NAS-S denote the corresponding method on the basic search space wile I-Supernet-L, P-Supernet-L and PAD-NAS-L represent the one on the large search space. 

To make a fair comparison, for the three methods above, all the experimental setups are exactly the same, e.g., the total training epochs, learning rate, and hyperparameters in the CES. The only difference between these three methods is the number of progressive search stages $M$.

\subsection{Comparison with State-of-the-art Methods}
\noindent\textbf{MobileNetV2-based Results.} 	The experimental results are shown in Table~\ref{result}. We compare our searched networks with SOTA efficient networks both designed automatically and manually. The primary metrics we care about are top-1 accuracy and mobile latency. If the latency is not available, the FLOPs is used as the secondary efficiency metric.

\setlength{\tabcolsep}{2pt}
\begin{table}
	\centering
	\caption{Comparison with the SOTA on ImageNet. The FLOPs and Latency are calculated with 224$\times$224 input.}
	\scalebox{0.78}{\begin{tabular}{llcccc}  
			\hline
			
			No.  & Model  & Params & FLOPs & Latency & Top-1 Acc (\(\% \)) \\
			\hline
			1 & MobileNetV1 \cite{howard2017mobilenets}                  & 4.2M  & 569M   &89.97ms  & 70.6     \\
			2 & MobileNetV2 \cite{sandler2018mobilenetv2}              & 3.4M  & 300M &61.13ms & 72.0     \\
			3 & MobileNetV2  (\( \times 1.4\))  \cite{sandler2018mobilenetv2}             & 6.9M  & 585M &104.65ms & 74.7  \\
			4 & ShuffleNetV2 \cite{ma2018shufflenet}                       & 3.4M  & 299M &-   & 72.6     \\
			\hline
			5 & NASNet-mobile \cite{zoph2018learning}                    &5.3M   & 564M & 144.05ms  &74.0   \\
			6 & DARTS \cite{liu2018darts}                                         &4.7M   & 574M  &-   & 73.3       \\
			7 & P-DARTS \cite{chen2019progressive}                         &4.9M   & 574M  &-   & 75.6   \\
			8 & SPOS \cite{guo2019single}                                        & 3.5M   & 319M  &-  & 74.3       \\
			9 & REGNETX-400MF \cite{radosavovic2020designing}    & 5.2M  & 400M &-  &72.7      \\
			10 & MnasNet \cite{tan2019mnasnet}                                &4.2M   & 317M &64.24ms   & 74.0    \\
			11 & DenseNAS-B \cite{fang2019densely}              & -  & 314M &66.62ms  &74.6      \\
			12 & ABS \cite{hu2020angle}                                & 3.8M  & 325M &68.21ms  &74.4      \\
			13 & Proxyless-R (mobile) \cite{cai2018proxylessnas}       & 4.1M  & 320M &67.66ms & 74.6    \\
			14 & FBNet-A \cite{wu2019fbnet}                                     & 4.3M  & 249M &54.05ms & 73.0  \\
			15 & FBNet-B \cite{wu2019fbnet}                                     & 4.5M  & 295M &63.79ms & 74.1    \\
			16 & FBNet-C \cite{wu2019fbnet}                                     & 5.5M  & 375M &79.18ms & 74.9   \\
			17 & Single Path NAS \cite{stamoulis2019single}              & 4.4M  & 334M &68.67ms  &74.9      \\
			18 & FairNAS-C \cite{chu2019fairnas}              & 4.4M  & 321M &67.32ms  &74.7     \\			
			\hline
			19 & PAD-NAS-S-A   &3.7M    &310M &61.31ms  &74.6        \\
			20 & PAD-NAS-S-B   &4.1M   &315M &64.71ms  &74.9      \\
			21 & PAD-NAS-S-C  &4.2M   &347M &69.85ms  &75.2        \\
			\hline
			22 & PAD-NAS-L-A   &4.0M   &271M &54.72ms  &74.5       \\
			23 & PAD-NAS-L-B   &4.2M   &334M &67.23ms  &75.2      \\
			24 & PAD-NAS-L-C  &4.6M   &385M &79.08ms  &75.7      \\
			25 & PAD-NAS-L-D  &4.6M   &401M &90.21ms  &75.9      \\
			26 & PAD-NAS-L-E  &4.7M   &444M &101.53ms  &\textbf{76.1}    \\
			\hline
	\end{tabular}}
	\label{result}
	\vspace{-0.5cm}
\end{table}

For the basic search space, we present three models PAD-NAS-S-A, PAD-NAS-S-B and PAD-NAS-S-C according to their latency. Compared to manually designed network, as shown the first four rows in Table~\ref{result}, it can reach significantly higher accuracy with lower latency(or FLOPs). In particular, PAD-NAS-S-C achieves ${75.2\%}$ with 69.85ms latency, which surpasses MobileNetV2(1.4X) top-1 accuracy by ${0.4\%}$ while much faster (from 104.65ms to 69.85ms). Comapared with Proxyless-R \cite{cai2018proxylessnas}  which shares exactly the same search space, ours PAD-NAS-S-B improves the accuracy by ${0.3\% }$ but still faster (2.95ms less for processing single image on Pixel 3). Finally, in comparison with other SOTA NAS algorithms, the three models all achieve higher accuracy with lower latency.

\begin{figure}
	\vspace{-0.2cm}
	\centering
	\includegraphics[scale=0.19]{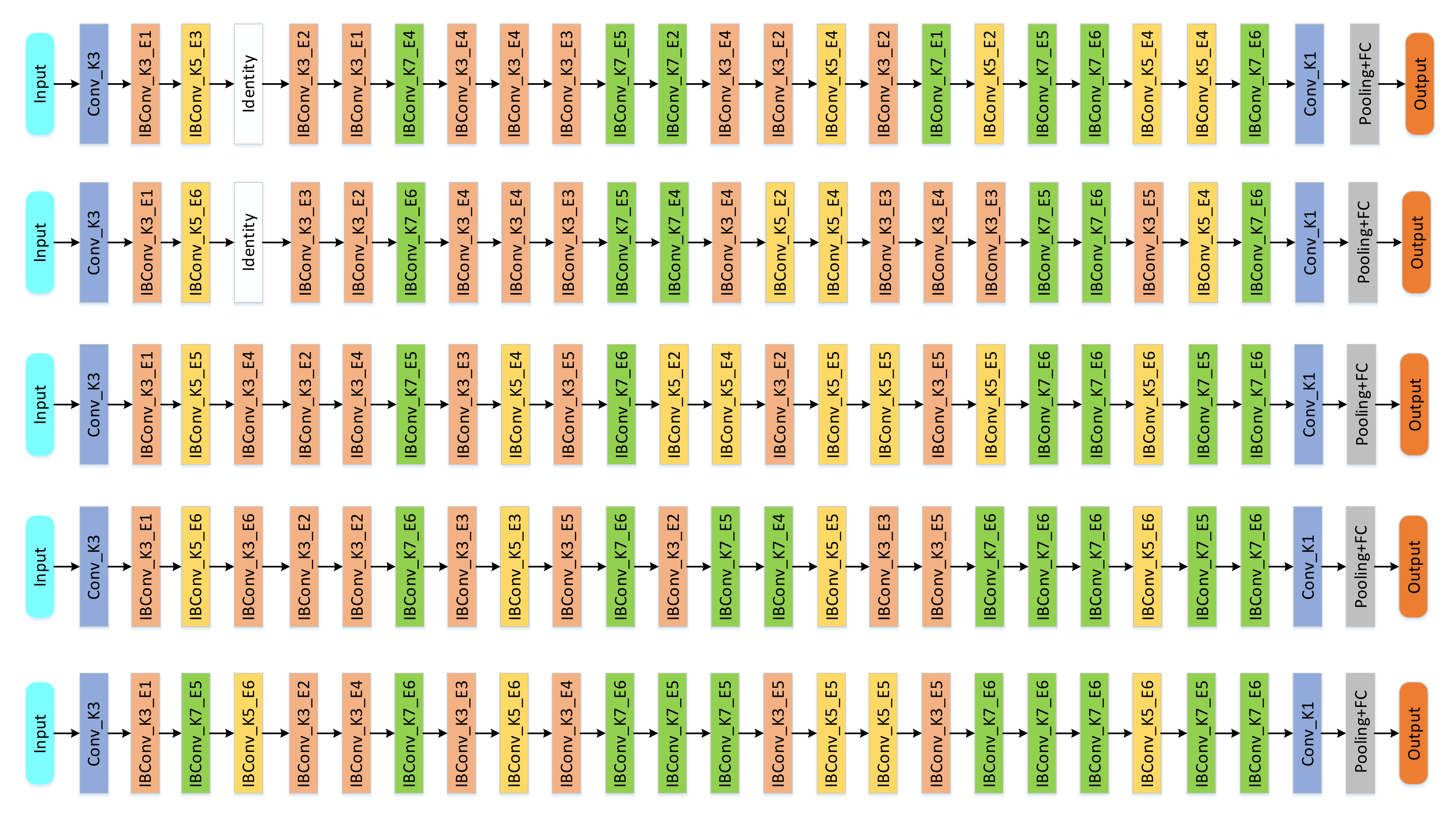}
	\vspace{-0.3cm}
	\caption{Architectures of PAD-NAS-L-A,B,C,D,E in Fig.~\ref{acc} (from top to bottom).}
	\label{searched}
	\vspace{-0.1cm}
\end{figure}

For the large search space, we list five models according to their latency, named PAD-NAS-L-A, PAD-NAS-L-B, PAD-NAS-L-C, PAD-NAS-L-D and PAD-NAS-L-E, whose detailed architectures are  illustrated in Fig.~\ref{searched}. The top-1 accuracy of PAD-NAS-L-A is ${74.5\%}$, which is ${1.5\%}$ higher than its counterpart FBNet-A \cite{wu2019fbnet}.
PAD-NAS-L-B achieves ${75.2\%}$ top-1 accuracy, which is ${0.6\%}$ higher than Proxyless-R \cite{cai2018proxylessnas}, and exhibits the same accuracy as PAD-NAS-S-C with lower latency. PAD-NAS-L-C achieves ${75.7\%}$ top-1 accuracy, ${0.8\%}$ higher than FBNet-C \cite{wu2019fbnet}.  These results demonstrate our proposed PAD-NAS can be applied to larger search space, and achieve even better performance. Moreover, PAD-NAS can search for multiple models with various latency from the refined search space(from PAD-NAS-L-A to PAD-NAS-L-E), which verifies the model diversity is maintained.

\noindent\textbf{MobileNetV3-based Results.} The experimental results with MobileNetV3-based search space (PAD-NAS-V3) are shown in Table~\ref{mv3}, where SE and swish are used. We can see PAD-NAS model achieves better performance.
\begin{table}[H]
	\vspace{-0.3cm}
	\centering
	\caption{Comparison with the SOTA under MobileNetV3-based search space.}
	\scalebox{0.8}{\begin{tabular}{l|ccc}  
			\hline
			Model & FLOPs   &Latency &Top-1 Acc (\(\% \))  \\
			\hline
			MobileNetV3 (\( \times 0.75\)) \cite{howard2019searching} & 155M    & 40.4ms  &73.3  \\
			MobileNetV3\cite{howard2019searching}  &219M    & 52.6ms &75.2 \\ 
			GreedyNAS-C\cite{you2020greedynas}  &284M    & 58.1ms &76.2 \\ 
			\hline
			PAD-NAS-V3-A & 182M   & 40.8ms  &75.0 \\
			PAD-NAS-V3-B &222M  & 49.2ms &75.9  \\
			PAD-NAS-V3-C &266M   & 55.1ms &\textbf{76.5}  \\
			\hline
	\end{tabular}}
	\label{mv3}
	\vspace{-0.6cm}
\end{table}

	\vspace{0.1cm}
\noindent\textbf{Transfer Learning Results.} We transfer the backbone searched by PAD-NAS to semantic segmentation task on cityscapes. The results are shown in Table~\ref{Segmentation}. The result for MobilenetV2 is reported in \cite{howard2019searching}. We can see PAD-NAS model outperforms efficient networks automatically and manually.

\begin{table}[H]
	\vspace{-0.4cm}
	\centering
	\caption{Semantic segmentation results on Cityscapes val set. The FLOPs and Latency are
		calculated with 512$\times$1024 input.}
	\scalebox{0.9}{\begin{tabular}{l|ccc}  
			\hline
			Model & FLOPs   &Latency &  mIOU(\(\% \)) \\
			\hline
			MobileNetV2 & 12.6B     & 773ms   &72.7   \\
			Proxyless-R (mobile) & 13.5B     & 831ms   &73.1   \\
			PAD-NAS-L-A & 11.3B   & 680ms  &73.3\\
			\hline
	\end{tabular}}
	\label{Segmentation}
	\vspace{-0.2cm}
\end{table}

\subsection{Automatically Designed Search Space}	
\noindent\textbf{Impact of Search Space.} Theoretically, NAS should give better result when larger search space is used. However, as stated in Table~\ref{supernet_comparsion}, I-Supernet-L achieves even lower stand-alone top-1 accuracy than I-Supernet-S. There exists a gap between theory and practice in current one-shot approaches. More operations brings more intense interference between operations during the training of supernet, which makes the weight coupling more severe. Thanks to the proposed progressive search strategy, PAD-NAS-L performs better than PAD-NAS-S, which implies PAD-NAS can bridge this gap. Besides, we can see top-1 accuracy achieved by PAD-NAS-L is 1.02$\%$ higher than I-Supernet-L while this number is only 0.42$\%$ for basic search space. This shows the advantage of PAD-NAS is more obvious in larger search space. We can also notice big gap exists between supernet accuracy and stand-alone one in the baseline method I-Supernet, as large as 21.99$\%$,  while this gap is much lower in PAD-NAS, only 2.94$\%$.
\begin{table}[H]
	\vspace{-0.3cm}
	\centering
	\caption{Performance comparison between I-Supernet and PAD-NAS on different search spaces. Stand-alone Top-1 Acc means the best top-1 accuracy of searched architecture when stand-alone trained from scratch under the same latency(69ms).}
	\scalebox{0.84}{\begin{tabular}{l|c|cc}  
			\hline
			Algorithm & Supernet Top-1 Acc (\(\% \))   &  Stand-alone Top-1 Acc (\(\% \)) \\
			\hline
			I-Supernet-L           & 52.32$\pm$0.51     & 74.31$\pm$0.14     \\
			PAD-NAS-L           & 72.39$\pm$0.08       & 75.33$\pm$0.09  \\
			I-Supernet-S        & 65.51$\pm$0.13          & 74.70$\pm$0.09    \\
			PAD-NAS-S           & 71.42$\pm$0.07         & 75.12$\pm$0.08 \\
			\hline
	\end{tabular}}
	\label{supernet_comparsion}
	\vspace{-0.5cm}
\end{table}
\begin{figure}[h]
	\vspace{-0.2cm}
	\centering
	\includegraphics[scale=0.23]{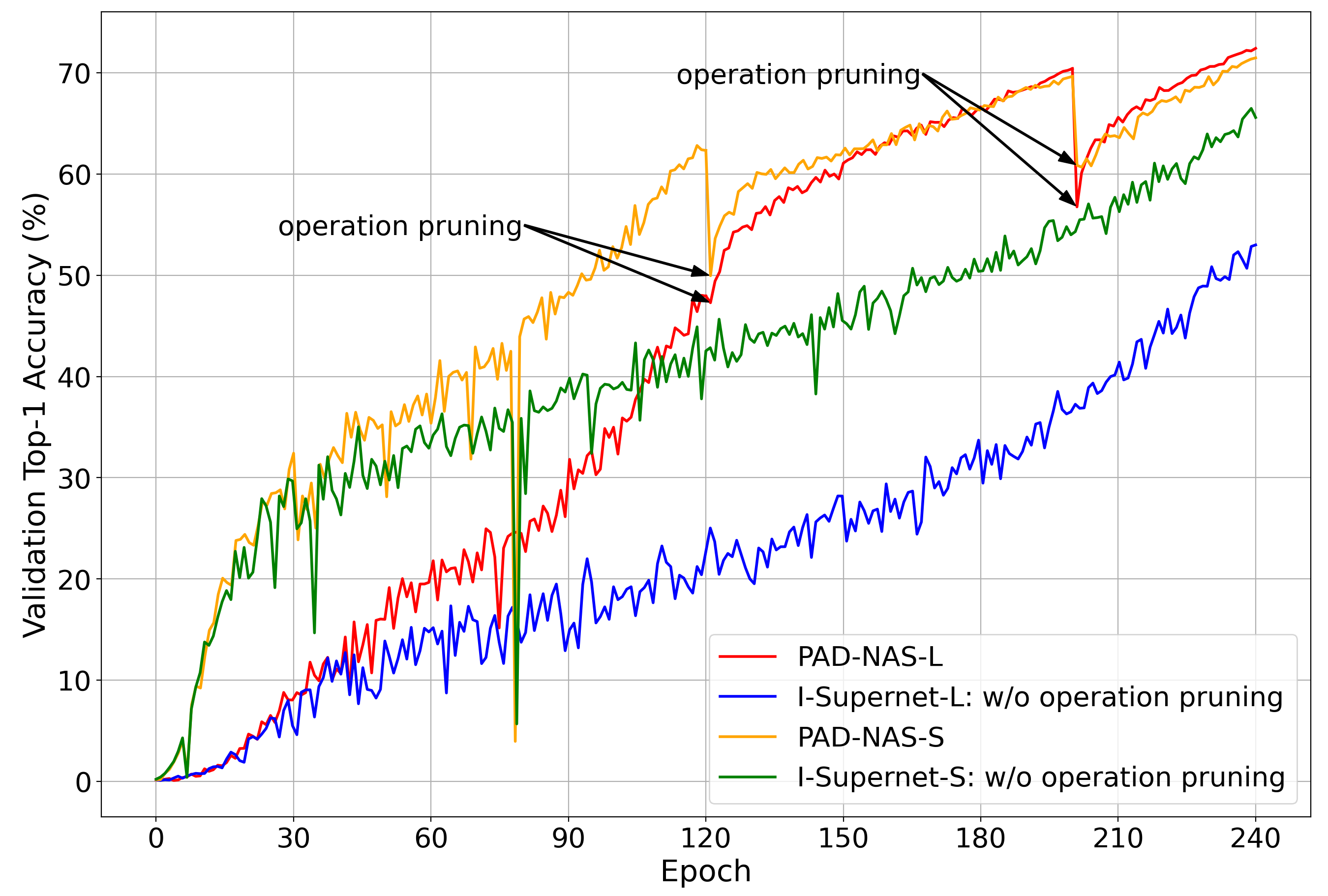}
	\caption{Supernet accuracy curve comparison between I-Supernet and PAD-NAS.}
	\label{traincurve}
	\vspace{-0.2cm}
\end{figure}

We also present the corresponding supernet accuracy curve on validation dataset for PAD-NAS and I-Supernet in Fig.~\ref{traincurve}.
The supernet accuracy of I-Supernet-L is always below I-Supernet-S. Similar trend also exists in the first 120 epochs of PAD-NAS. However, once operation pruning is introduced, the gap between PAD-NAS-L and PAD-NAS-S is becoming much smaller. And  PAD-NAS-L eventually surpasses PAD-NAS-S at epoch 240 after we prune the operations twice. Here, we repeat the experiments 5 times with different random seeds.

\noindent\textbf{Random Search Baseline.} Under the same latency constrain,  we randomly select 5 models from the basic small and large search space and denote their results as Random-S and Random-L. Moreover, we also randomly select 5 models from the Auto-Designed search space (AD) by PAD-NAS, whose results are named as Random-S-AD and Random-L-AD.
\begin{table}[H]
	\vspace{-0.2cm}
	\centering
	\caption{Random search results under the same latency(69ms).}
	\scalebox{1.0}{\begin{tabular}{l|c}  
			\hline
			Algorithms       & Top-1 Acc($\%$) \\
			\hline
			Random-L         & 72.71$\pm$0.46      \\
			Random-L-AD        & 74.91$\pm$0.11   \\		
			\hline	
			Random-S        & 73.88$\pm$0.19  \\
			Random-S-AD     & 74.72$\pm$0.10   \\	
			\hline
	\end{tabular}}
	\label{random_search}
	\vspace{-0.4cm}
\end{table}

As shown in Table~\ref{random_search}, the random search results corresponding to AD are greatly improved compared to the basic small search space(74.91$\%$ vs. 72.71$\%$ and 74.72$\%$ vs. 73.88$\%$). Even random search performs pretty well in the search space designed by PAD-NAS, which verifies the search space quality is improved.

\begin{figure}[h]
	\vspace{-0.3cm}
	\centering
	\includegraphics[scale=0.35]{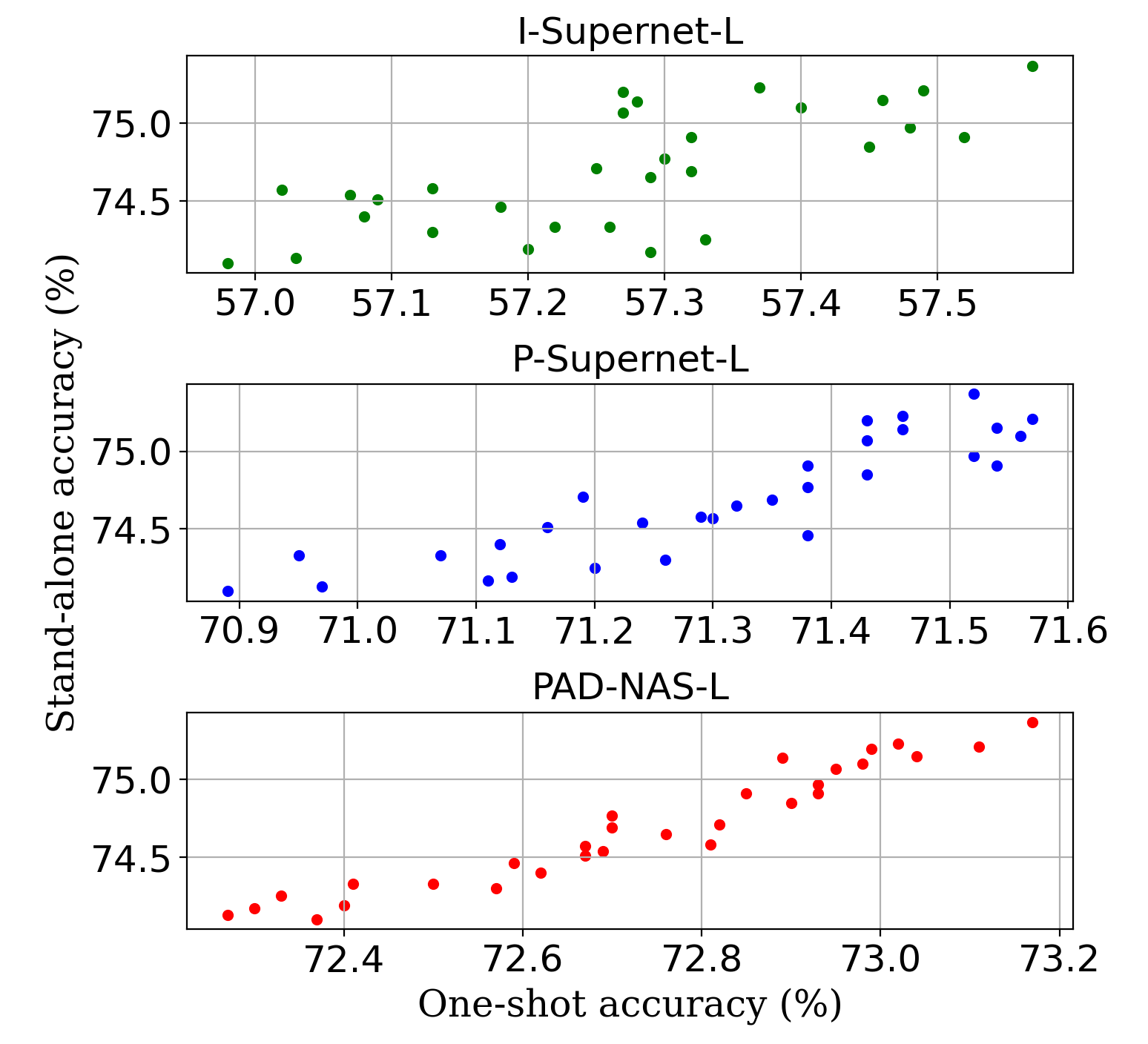}
	\caption{Top-1 accuracy of 30 stand-alone trained architectures vs. one-shot models.}
	\label{ranking}
	\vspace{-0.3cm}
\end{figure}

\vspace{0.1cm}
\noindent\textbf{Kendall Rank Analysis.} The ranking consistency of stand-alone and one-shot model accuracy is an important problem in One-Shot NAS algorithm. It indicates whether supernet can reflect the relative performance of models.
Due to high training cost, we sample 30 models at approximately equal distances on the Pareto frontier and do the stand-alone training from scratch to get the ranking, as shown in Fig.~\ref{ranking}. We can observe in PAD-NAS-L, one-shot accuracy is more relevant to stand-alone accuracy, compared to the baseline method I-Supernet-L where there is no operation pruning involved. 
\begin{table}[H]
		\vspace{-0.2cm}
	\centering
	\caption{Ranking consistency comparison with baseline methods. Size represents search space size.}
	\begin{tabular}{lccccccccc}  
		\hline
		Algorithm &  &  & $\tau$ & & &  Size  & & & M   \\
		\hline
		I-Supernet-S &  & &  0.6153$\pm$0.0387 & &  & $ {10^{17}} $ & & & 2     \\
		I-Supernet-L &  & &  0.4977$\pm$0.0573  & &  & $ {10^{27}} $ & & & 2     \\
		P-Supernet-L &  & & 0.7451$\pm$0.0332  & & & $ {10^{15}} $ & & & 3 \\
		PAD-NAS-L & & &  0.8879$\pm$0.0201  & & & $ { 10^{9} }$ & & &  4      \\
		\hline
	\end{tabular}
	\label{rank}
	\vspace{-0.4cm}
\end{table}

We also use Kendall's rank correlation coefficient $\tau$ \cite{kendall1938new} to measure this consistency. $\tau$ ranges from -1 to 1, meaning the rankings are totally reversed or completely preserved, whereas 0 means there is no correlation at all. As shown in Table~\ref{rank}, ${\tau}$ for I-Supernet-L is only 0.4977, while the corresponding value for I-Supernet-S is 0.6153. Thanks to the proposed progressive search strategy, as the number of stage $M$ increases, the spaces size is reduced dramatically by orders-of-magnitude, and Kendall's ranking coefficient $\tau$ has significantly increased from 0.4977 to 0.8879, which indicates weight coupling is alleviated. Here, we repeat the experiments 3 times with different random seeds.

\vspace{-0.1cm}

\subsection{Ablation Studies}

\vspace{-0.2cm}
\noindent\textbf{ Search Cost Analysis.} Evolution search is performed twice more than the baseline method, which costs 7 more hours. All experiments are trained with 8 Tesla V100 GPUs. The analysis is shown below in Table~\ref{search_cost}.
\begin{table}[H]
	\vspace{-0.1cm}
	\centering
	\caption{The comparison of search cost.}
	\scalebox{0.95}{\begin{tabular}{l|cc}  
			\hline
			Method & SPOS\cite{guo2019single}   &PAD-NAS  \\
			\hline
			Supernet training & 32hours     & 39hours    \\
			Evolution search &3.5hours    & 3.5hours \\
			Retrain & 24hours   & 24hours  \\	\hline
			Total &59.5hours   & 66.5hours  \\
			\hline
	\end{tabular}}
	\label{search_cost}
	\vspace{-0.5cm}
\end{table}

\begin{figure}[h]
	
	\centering
	\includegraphics[scale=0.32]{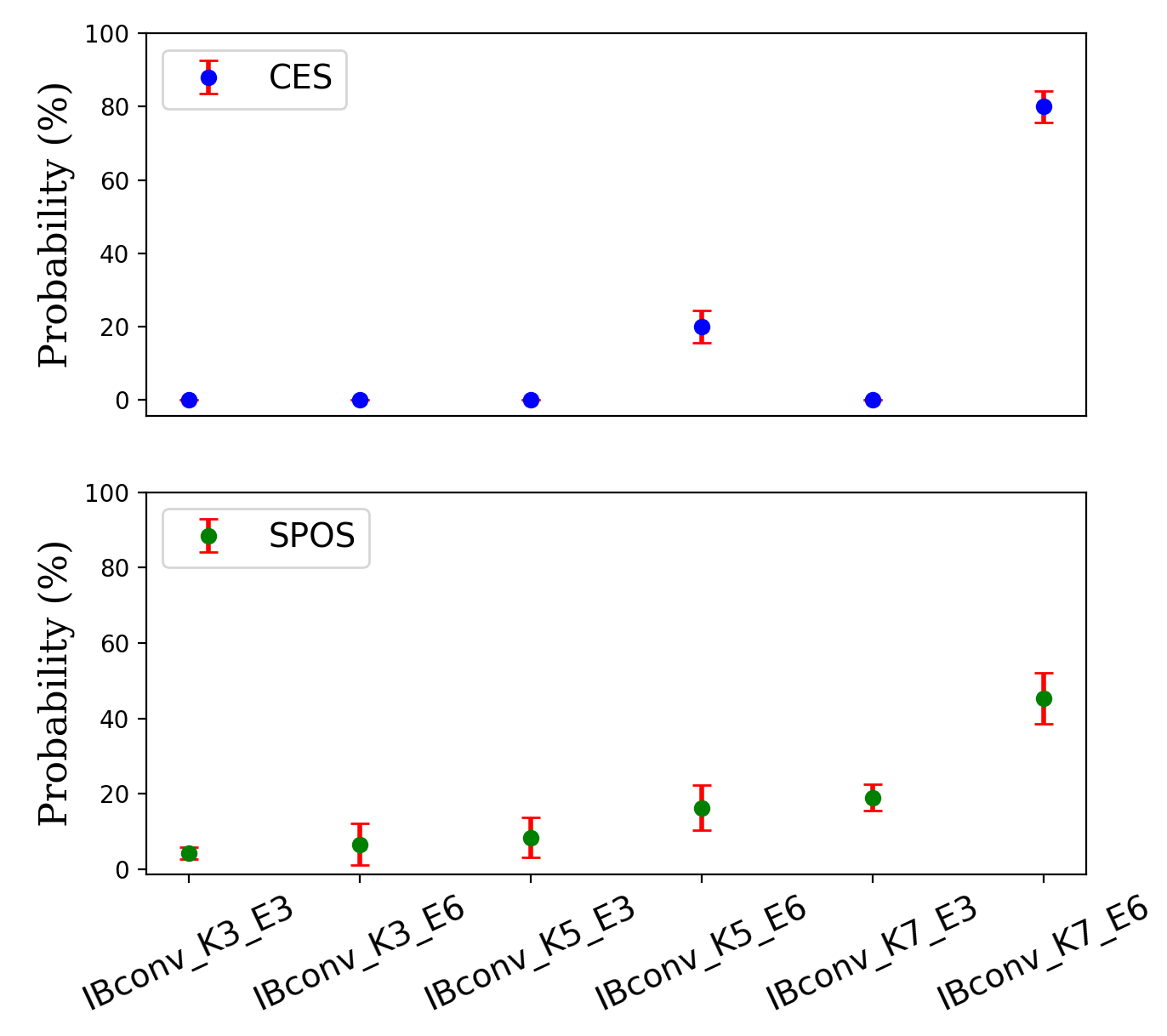}
	\caption{Estimated probability distribution of the operations in the last layer  by CES and SPOS \cite{guo2019single}, respectively.}
	\label{eaprob}
	\vspace{-0.4cm}
\end{figure}	

\noindent\textbf{Impact of Evolutionary Algorithm.} We study the impact of different evolutionary algorithms on the estimated probability distribution of operations. Two algorithms are compared here: one is our CES while the other is the one used in SPOS \cite{guo2019single}. Each algorithm is repeated 5 times with different random seeds to get the mean and variance of estimated distribution. As illustrated in Fig.~\ref{eaprob}, the result of our CES exhibits a much lower variance than SPOS \cite{guo2019single}, which indicates CES gives us more stable result. Besides, the distribution corresponding to CES is sparser than SPOS, which makes operation pruning more efficient.



%

\section{Conclusion}
In this paper, we present PAD-NAS, a progressive search strategy for One-Shot NAS. It automatically designs a layer-wise operation search space through operation pruning. PAD-NAS mitigates weight coupling issue and significantly improves the ranking consistency between supernet predicted accuracy and stand-alone trained accuracy. Experimental results demonstrate the effectiveness of our method,
which achieves SOTA performance on ImageNet.

{\small
\bibliographystyle{ieee_fullname}
\bibliography{egbib}
}

\end{document}